\documentclass[11pt]{article}
\usepackage{acl2016}
\usepackage{main}
\usepackage{times}
\usepackage{lipsum}
\usepackage{latexsym}
\usepackage{booktabs}
\usepackage[normalem]{ulem}
\useunder{\uline}{\ul}{}

\aclfinalcopy % Uncomment this line for the final submission
\DeclareMathSizes{4}{4}{4}{4}
\DeclareMathSizes{3}{17.28}{9}{7} 

\newcommand{\bea}{\begin{eqnarray}}
\newcommand{\eea}{\end{eqnarray}}

\title{Pointing the Unknown Words}
\author{Caglar Gulcehre \\
  Universit\'e de Montr\'eal \\\And 
  Sungjin Ahn \\\ 
  Universit\'e de Montr\'eal \\\And 
  Ramesh Nallapati \\
  IBM T.J. Watson Research \\\AND 
  Bowen Zhou \\
  IBM T.J. Watson Research \\\And 
  Yoshua Bengio\\
  Universit\'e de Montr\'eal\\
  CIFAR Senior Fellow\\ \\}
 
\date{}

\begin{document}

\maketitle

\begin{abstract}
The problem of rare and unknown words is an important issue that can potentially effect the
performance of many NLP systems, including both the traditional count-based and the deep learning
models. We propose a novel way to deal with the rare and unseen words for the neural network
models using attention. Our model uses two softmax layers in order to predict the next word in
conditional language models: one predicts the location of a word in the source sentence, and the
other predicts a word in the shortlist vocabulary. At each time-step, the decision of which
softmax layer to use choose adaptively made by an MLP which is conditioned on the context.~We
motivate our work from a psychological evidence that humans naturally have a tendency to point
towards objects in the context or the environment when the name of an object is not known.~We
observe improvements on two tasks, neural machine translation on the Europarl English to French
parallel corpora and text summarization on the Gigaword dataset using our proposed model.
\end{abstract}

\section{Introduction}

Words are the basic input/output units in most of the NLP systems, and thus the ability to cover a
large number of words is a key to building a robust NLP system. However, considering that (i) the
number of all words in a language including named entities is very large and that (ii) language
itself is an evolving system (people create new words), this can be a challenging problem.

A common approach followed by the recent neural network based NLP systems is to use a softmax
output layer where each of the output dimension corresponds to a word in a predefined
word-shortlist.~Because computing high dimensional softmax is computationally expensive, in
practice the shortlist is limited to have only top-$K$ most frequent words in the training corpus.
All other words are then replaced by a special word, called the \textit{unknown word (UNK)}.

The shortlist approach has two fundamental problems. The first problem, which is known as the
\textit{rare word} problem, is that some of the words in the shortlist occur less frequently in
the training set and thus are difficult to learn a good representation, resulting in poor
performance. Second, it is obvious that we can lose some important information by mapping
different words to a single dummy token \textit{UNK}. Even if we have a very large shortlist
including all unique words in the training set, it does not necessarily improve the test
performance, because there still exists a chance to see an unknown word at test time. This is
known as the \textit{unknown word} problem. In addition, increasing the shortlist size mostly
leads to increasing rare words due to Zipf's Law. 

These two problems can be particularly critical in language understanding tasks such as factoid
question answering \cite{bordes2015large} where the words that we are interested in are often
named entities which are usually unknown or rare words. 

In a similar situation, where we have a limited information on how to call an object of interest,
   it seems that humans (and also some primates) have an efficient behavioral mechanism of drawing
   attention to the object: \textit{pointing} \cite{matthews2012origins}.  Pointing makes it
   possible to deliver information and to associate context to a particular object without knowing
   how to call it. In particular, human infants use pointing as a fundamental communication tool
   \cite{tomasello2007new}. 

\begin{figure}[ht]
    \includegraphics[width=0.95\columnwidth]{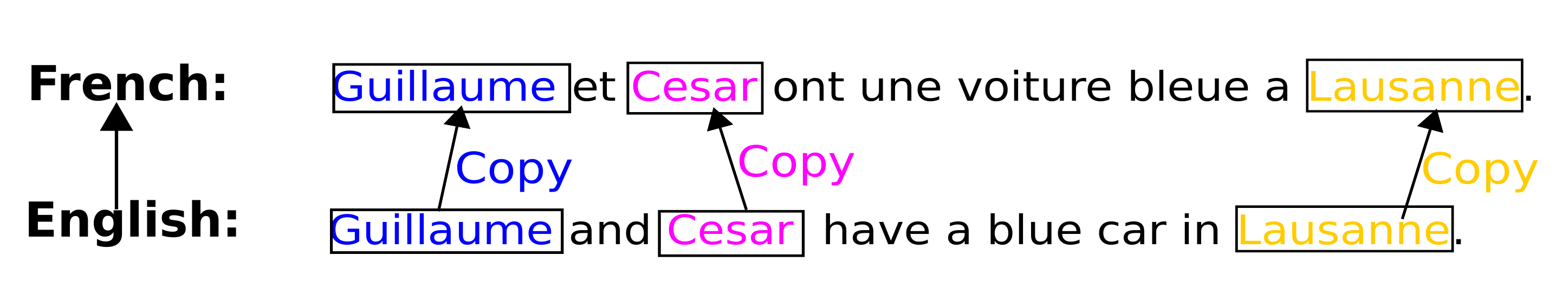}
    \caption{An example of how copying can happen for machine translation. Common words that appear both in source and the target can directly be copied from input to source. The rest of the unknown in the target can be copied from the input after being translated with a dictionary.}
    \label{fig:nmt_word_copying_ex}
    \centering
\end{figure}

In this paper, inspired by the pointing behavior of humans and recent advances in the attention
mechanism \cite{nmt} and the pointer networks \cite{vinyals2015pointer}, we propose a novel method
to deal with the rare or unknown word problem. The basic idea is that we can see in many NLP
problems as a task of predicting target text given context text, where some of the target words
appear in the context as well. We observe that in this case we can make the model \textit{learn to
point} a word in the context and copy it to the target text, as well as \textit{when to
point}. For example, in machine translation, we can see the source sentence as the
context, and the target sentence as what we need to predict. In Figure
\ref{fig:nmt_word_copying_ex}, we show an example depiction of how words can be copied
from source to target in machine translation. Although the source and target languages are
different, many of the words such as named entities are usually represented by the same
characters in both languages, making it possible to copy. Similarly, in text
summarization, it is natural to use some words in the original text in the summarized text
as well. 

Specifically, to predict a target word at each timestep, our model first determines  the source of
the word generation, that is, on whether to take one from a predefined  shortlist or to copy one from
the context. For the former, we apply the typical softmax operation, and for the latter, we use
the attention mechanism to obtain the pointing softmax probability over the context words and pick
the one of high probability. The model learns this decision so as to use the pointing only when
the context includes a word that can be copied to the target.~This way, our model can predict even
the words which are not in the shortlist, as long as it appears in the context. Although some of
the words still need to be labeled as \textit{UNK}, i.e., if it is neither in the shortlist nor in
the context, in experiments we show that this \textit{learning when and where to point} improves
the performance in machine translation and text summarization.

The rest of the paper is organized as follows. In the next section, we review the related works 
including pointer networks and previous approaches to the rare/unknown problem. In Section 3, 
we review the neural machine translation with attention mechanism which is the baseline in 
our experiments. Then, in Section 4, we propose our method dealing with the rare/unknown 
word problem, called the {\bf Pointer Softmax~(PS)}. The experimental results are provided in the Section 5 and we conclude our work in Section 6.

\section{Related Work}

The attention-based pointing mechanism is introduced first in the pointer networks \cite{vinyals2015pointer}. 
In the pointer networks, the output space of the target sequence is constrained to be the observations in 
the input sequence (not the input space). Instead of having a fixed dimension softmax output layer, 
softmax outputs of varying dimension is dynamically computed for each input sequence 
in such a way to maximize the attention probability of the target input. However, its 
applicability is rather limited because, unlike our model, there is no option to choose 
whether to point or not; it always points. In this sense, we can see the pointer networks 
as a special case of our model where we always choose to point a context word.

Several approaches have been proposed towards solving the rare words/unknown words 
problem, which can be broadly divided into three categories. The first category of 
the approaches focuses on improving the computation speed of the softmax output so that 
it can maintain a very large vocabulary. Because this only increases 
the shortlist size, it helps to mitigate the unknown word problem, but still suffers 
from the rare word problem. The hierarchical softmax \cite{hiersoftmax}, importance sampling \cite{bengio2008adaptive,Jean2014}, and the noise contrastive estimation \cite{nce,mnih2013learning} methods are in the class. 
% of algorithms that are trying to address either the issues regarding to unknown-words or scalability of the large-softmax. 

The second category, where our proposed method also belongs to, uses information from 
the context. Notable works are \cite{luong2015addressing} and \cite{hermann2015teaching}. In particular, 
applying to machine translation task, \cite{luong2015addressing} learns to point some words in 
source sentence and copy it to the target sentence, similarly to our method. However, it does not use 
attention mechanism, and by having fixed sized softmax output over the relative pointing range 
(e.g., -7, \dots, -1, 0, 1, \dots, 7), their model (the Positional All model) has a limitation 
in applying to more general problems such as summarization and question answering, where, unlike 
machine translation, the length of the context and the pointing locations in the context can vary dramatically. In question answering setting, \cite{hermann2015teaching} have used placeholders on named entities in the context. However, the placeholder id is directly predicted in the softmax output rather than predicting its location in the context. 
% Note that the first and second classes can be applied jointly. 

The third category of the approaches changes the unit of input/output itself from 
words to a smaller resolution such as characters \cite{graves2013generating} or bytecodes 
\cite{sennrich2015neural,gillick2015multilingual}. Although this approach has the main 
advantage that it could suffer less from the rare/unknown word problem, the training 
usually becomes much harder because the length of sequences significantly increases. 

Simultaneously to our work, \cite{gu2016incorporating} and \cite{cheng2016neural} proposed models that learn to copy from source to target and both papers analyzed their models on summarization tasks.  

\section{Neural Machine Translation Model with Attention}
\label{sec:model_des}
As the baseline neural machine translation system, we use the model proposed by \cite{nmt} that learns to (soft-)align and translate jointly. We refer this model as NMT.

The encoder of the NMT is a bidirectional RNN~\cite{Schuster1997}. The forward RNN reads input
sequence $\vx=(x_1, \dots, x_T)$ in left-to-right direction, resulting in a 
sequence of hidden states $(\ora{\vh}_1, \dots, \ora{\vh}_T)$. The backward RNN
reads $\vx$ in the reversed direction and outputs $(\ola{\vh}_1, \dots,
\ola{\vh}_T)$. We then concatenate the hidden states of forward and backward RNNs 
at each time step and obtain a sequence of {\it annotation} vectors $(\vh_1, \dots, \vh_T)$ 
where $\vh_j = \left[\ora{\vh}_j || \ola{\vh}_j\right]$. Here, $||$ denotes the concatenation 
operator. Thus, each annotation vector $\vh_j$ encodes information about the $j$-th word with
respect to all the other surrounding words in both directions.

In the decoder, we usually use gated recurrent unit (GRU) \cite{cho2014learning,gru_rnn}. 
Specifically, at each time-step $t$, the soft-alignment mechanism first computes 
the relevance weight $e_{tj}$ which determines the contribution of annotation vector 
$\vh_j$ to the $t$-th target word. We use a non-linear mapping $f$ (e.g., MLP) 
which takes $\vh_j$, the previous decoder's hidden state $\vs_{t-1}$ and 
the previous output $y_{t-1}$ as input:
\begin{align*}
    e_{tj} = f(\vs_{t-1}, \vh_j, y_{t-1}).
\end{align*}

The outputs $e_{tj}$ are then normalized as follows:
\begin{equation}
    \label{eq:dec_alpha}
    l_{tj} = \frac{\text{exp}(e_{tj})}{\sum_{k=1}^T\text{exp}(e_{tk})}.
\end{equation}

We call $\l_{tj}$ as the relevance score, or the alignment weight, of the $j$-th annotation vector.

The relevance scores are used to get the {\it context vector} $\vc_t$ of the $t$-th 
target word in the translation:
\begin{align*}
    \vc_t = \sum_{j=1}^T l_{tj} \vh_j ~,
\end{align*}

The hidden state of the decoder $\vs_t$ is computed
based on the previous hidden state $\vs_{t-1}$, the context vector
$\vc_t$ and the output word of the previous time-step $y_{t-1}$:
\begin{equation}
    \label{eq:dec:hidden}
    \vs_t = f_r(\vs_{t-1}, y_{t-1}, \vc_t),
\end{equation} where $f_r$ is GRU.

We use a deep output layer~\cite{pascanu2013construct} to compute the conditional
distribution over words: 
\begin{equation}
    \begin{split}
    \label{eq:dec:output}
    p(&y_t = a | y_{<t}, \vx) \propto \\
    & \exp\left(\psi_{(\mW_o,\vb_o)}^a
    f_o(\vs_t, y_{t-1}, \vc_t)\right),
    \end{split}
\end{equation}
where $\mW$ is a learned weight matrix and $\vb$ is a bias of the output layer. 
$f_o$ is a single-layer feed-forward neural network. $\psi_{(\mW_o,\vb_o)}(\cdot)$ 
is a function that performs an affine transformation on its input. And the superscript 
$a$ in $\psi^a$ indicates the $a$-th column vector of $\psi$.

\begin{figure}[ht]
    \includegraphics[width=0.99\columnwidth]{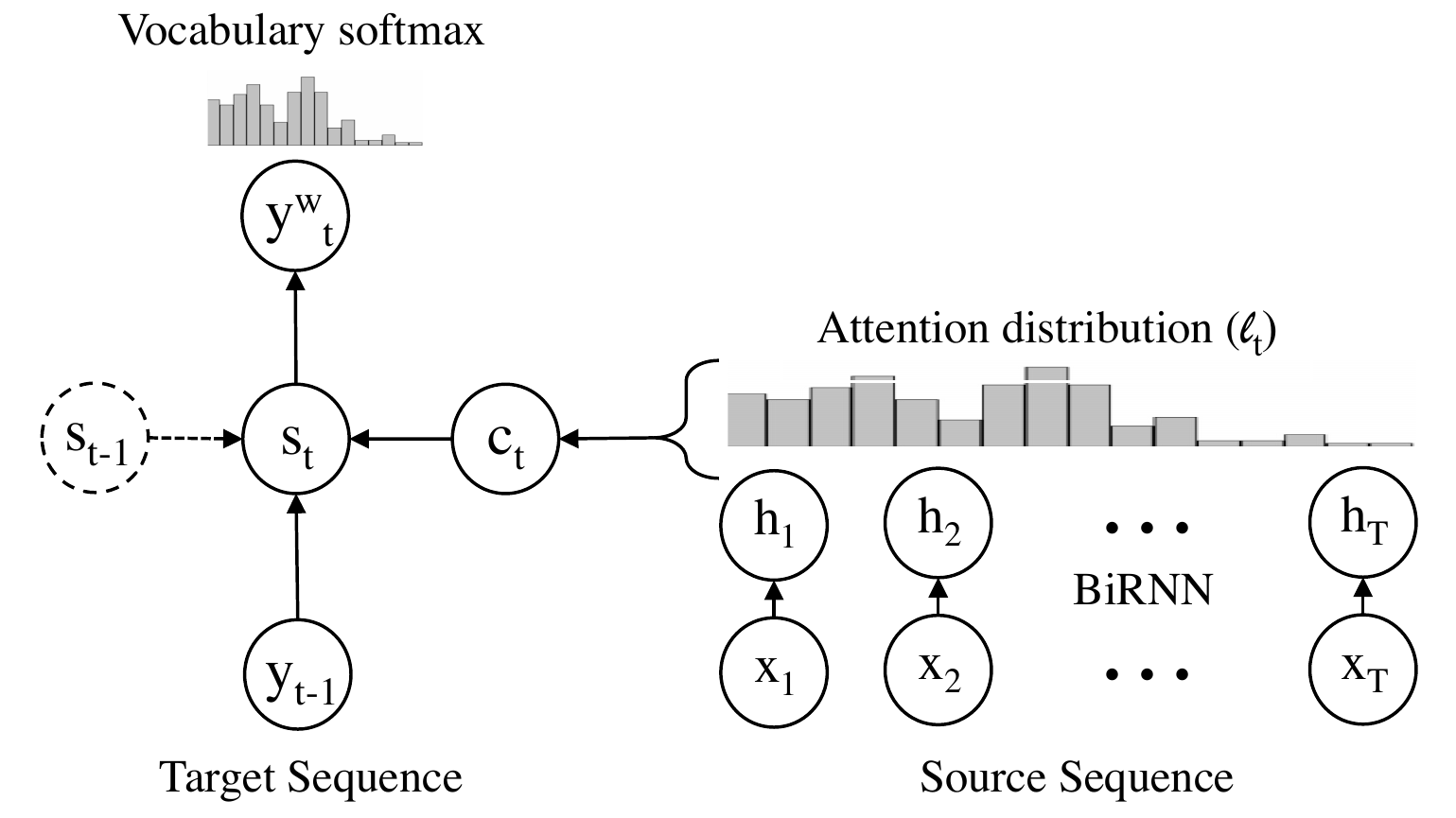}
    \caption{A depiction of neural machine translation architecture with attention. At each timestep, the model generates the attention distribution $\vl_t$. We use $\vl_t$ and the encoder's hidden states to obtain the context $\vc_t$. The decoder uses $\vc_t$ to predict a vector of probabilities for the words $\vw_t$ by using vocabulary softmax.}
    \label{fig:basic_att_model}
    \centering
\end{figure}

The whole model, including both the encoder and the decoder, is jointly trained to
maximize the (conditional) log-likelihood of target sequences given input 
sequences, where the training corpus is a set of $(\vx_n, \vy_n)$'s. Figure 
\ref{fig:basic_att_model} illustrates the architecture of the NMT.

\section{The Pointer Softmax}

In this section, we introduce our method, called as the pointer softmax (PS), to deal with the rare and unknown words. The pointer softmax can be an applicable approach to many NLP tasks, because it resolves the limitations about unknown words for neural networks. It can be used in parallel with other existing techniques such as the large vocabulary trick \cite{Jean2014}. Our model learns two key abilities jointly to make the pointing mechanism applicable in more general settings: (i) to predict whether it is required to use the pointing or not at each time step and (ii) to point any location of the context sequence whose length can vary widely over examples. Note that the pointer networks \cite{vinyals2015pointer} are in lack of the ability (i), and the ability (ii) is not achieved in the models by \cite{luong2015addressing}.

To achieve this, our model uses two softmax output layers, the \textit{shortlist} softmax and the \textit{location} softmax. The shortlist softmax is the same as the typical softmax output layer where each dimension corresponds a word in the predefined word shortlist. The location softmax is a pointer network where each of the output dimension corresponds to the location of a word in the context sequence. Thus, the output dimension of the location softmax varies according to the length of the given context sequence. 

\begin{figure}[ht]
    \includegraphics[width=0.99\columnwidth]{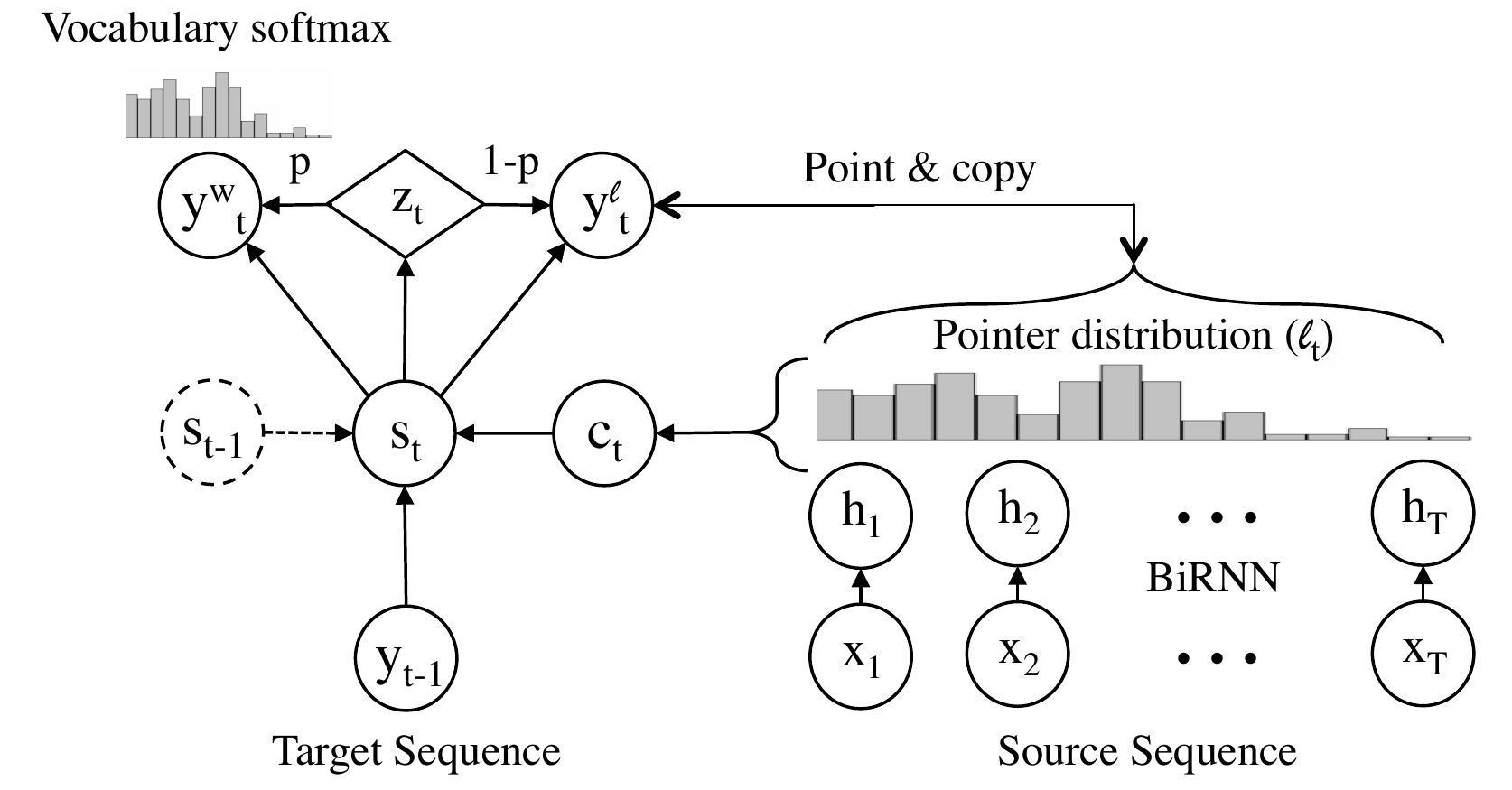}
    \caption{A depiction of the Pointer Softmax (PS) architecture. At each timestep,  $\vl_t$, $\vc_t$ and $\vw_t$ for the words over the limited vocabulary (shortlist) is generated. We have an additional switching variable $z_t$ that decides whether to use vocabulary word or to copy a word from the source sequence.}
    \label{fig:pointer_softmax}
    \centering
\end{figure}

At each time-step, if the model decides to use the shortlist softmax, we generate a word $w_t$ from the shortlist. Otherwise, if it is expected that the context sequence contains a word which needs to be generated at the time step, we obtain the location of the context word $l_t$ from the location softmax. The key to making this possible is deciding when to use the shortlist softmax or the location softmax at each time step. In order to accomplish this, we introduce a switching network to the model. The switching network, which is a multilayer perceptron in our experiments, takes the representation of the context sequence (similar to the input annotation in NMT) and the previous hidden state of the output RNN as its input. It outputs a binary variable $z_t$ which indicates whether to use the shortlist softmax (when $z_t = 1$) or the location softmax (when $z_t = 0$). Note that if the word that is expected to be generated at each time-step is neither in the shortlist nor in the context sequence, the switching network selects the shortlist softmax, and then the shortlist softmax predicts \textit{UNK}. The details of the pointer softmax model can be seen in Figure \ref{fig:pointer_softmax} as well.

More specifically, our goal is to maximize the probability of observing the target word 
sequence $\vy = (y_1, y_2, \dots, y_{T_y})$ and the word generation source $\vz = (z_1, z_2, \dots, z_{T_y})$, 
given the context sequence $\vx = (x_1, x_2, \dots, x_{T_x})$:
\bea
p_\theta(\vy,\vz|\vx) = \prod_{t=1}^{T_y}p_\theta(y_t, z_t| y_{<t},z_{<t}, \vx).
\eea
Note that the word observation $y_t$ can be either a word $w_t$ from the shortlist 
softmax or a location $l_t$ from the location softmax, depending on the switching variable $z_t$.

Considering this, we can factorize the above equation further 
\bea
p(\vy,\vz|\vx) &=& \prod_{t \in {\cal T}_w}p(w_t, z_t| (y,z)_{<t}, \vx)\times\nonumber\\
&&\prod_{t' \in {\cal T}_l}p(l_{t'}, z_{t'}| 
(y,z)_{<{t'}}, \vx).
\eea
Here, ${\cal T}_w$ is a set of time steps where $z_t = 1$, and ${\cal T}_l$ is a set of 
time-steps where $z_t = 0$. And, ${\cal T}_w \cup {\cal T}_l = \{1,2,\dots,T_y\}$ and 
${\cal T}_w \cap {\cal T}_l = \emptyset$. We denote all previous observations at step $t$ by $(y,z)_{<t}$. Note also that  $\vh_t = f((y,z)_{<t})$.

Then, the joint probabilities inside each product can be further factorized as follows:
\bea
p(w_t, z_t| (y,z)_{<t}) &=& p(w_t| z_t = 1, (y,z)_{<t})\times\nonumber\\ 
&&p(z_t = 1| (y,z)_{<t})\label{eqn:a}
\eea
\bea
p(l_t, z_t| (y,z)_{<t}) &=& p(l_t| z_t = 0, (y,z)_{<t})\times\nonumber\\ 
&&p(z_t = 0| (y,z)_{<t})\label{eqn:b}
\eea
here, we omitted $\vx$ which is conditioned on all probabilities in the above.

The switch probability is modeled as a multilayer perceptron with binary output:
\bea
p(z_t = 1| (y,z)_{<t}, \vx)~=~\sigmoid (f(\vx, \vh_{t-1}; \theta))\\
p(z_t = 0| (y,z)_{<t}, \vx)~=~1 - \sigmoid (f(\vx, \vh_{t-1}; \theta)).
\eea
And $p(w_t| z_t = 1, (y,z)_{<t}, \vx)$ is the shortlist softmax and $p(l_t| z_t = 0, (y,z)_{<t}, \vx)$ 
is the location softmax which can be a pointer network. $\sigmoid(\cdot)$ stands for the sigmoid function, $\sigmoid(x) = \frac{1}{\text{exp(-x)} + 1}$.

Given $N$ such context and target sequence pairs, our training objective is to maximize the following log likelihood w.r.t. the model parameter $\theta$
\bea
\argmax_\theta \frac{1}{N}\sum_{n=1}^N \log p_\theta(y_n,z_n|x_n).
\eea

\subsection{Basic Components of the Pointer Softmax}
In this section, we  discuss practical details of the three fundamental components of the pointer softmax. The interactions between these components and the model is depicted in Figure \ref{fig:pointer_softmax}.

\paragraph{Location Softmax $\vl_t$}: The location of the word to copy from source 
text to the target is predicted by the location softmax $\vl_t$. The location softmax outputs the conditional probability distribution $p(l_t| z_t = 0, (y,z)_{<t}, \vx)$. For models using the attention mechanism such as NMT, we can reuse the probability distributions over the source words in order to predict the location of the word to point. Otherwise we can simply use a pointer network of the model to predict the location.

\paragraph{Shortlist Softmax $\vw_t$} : The subset of the words in the vocabulary $V$ 
is being predicted by the shortlist softmax $\vw_t$. 

\paragraph{Switching network $d_t$}: The switching network $d_t$ is an MLP with sigmoid output function that outputs a scalar probability of switching between $\vl_t$ and $\vw_t$, and represents the conditional probability distribution $~p(z_t | (y,z)_{<t}, \vx)$. For NMT model, we condition the MLP that outputs the switching probability on the representation of the context of the source text $\vc_t$ and the hidden state of the decoder $\vh_t$. Note that, during the training, $d_t$ is observed, and thus we do not have to sample.

The output of the pointer softmax, $\vf_t$ will be the concatenation of the the two vectors, $d_t \times \vw_t$  and $(1-d_t)\times \vl_t$.

At test time, we compute Eqn. \eqref{eqn:a} and \eqref{eqn:b} for all shortlist word $w_t$ and all location $l_t$, and pick the word or location of the highest probability.

\section{Experiments}
In this section, we provide our main experimental results with the pointer softmax on machine translation and summarization tasks. In our experiments, we have used the same baseline model and just replaced the softmax layer with pointer softmax layer at the language model. We use the Adadelta~\cite{zeiler2012adadelta} learning rule for the training of NMT models. The code for pointer softmax model is available at \url{https://github.com/caglar/pointer_softmax}.

%\subsection{A toy QA-task}
%We constructed a toy task, in order to understand the efficiency of the PS.
\subsection{The Rarest Word Detection}
We construct a synthetic task and run some preliminary experiments in order to compare the results with the pointer softmax and the regular softmax's performance for the rare-words. The vocabulary size of our synthetic task is $|V|=600$ using sequences of length $7$. The words in the sequences are sampled according to their unigram distribution which has the form of a geometric distribution. The task is to predict the least frequent word in the sequence according to unigram distribution of the words. During the training, the sequences are generated randomly. Before the training, validation and test sets are constructed with a fixed seed.

We use a GRU layer over the input sequence and take the last-hidden state, in order to get the summary $\vc_t$ of the input sequence. The $\vw_t,~\vl_t$ are only conditioned on $\vc_t$, and the MLP predicting the $d_t$ is conditioned on the latent representations of $\vw_t$ and $\vl_t$. We use minibatches of size 250 using adam adaptive learning rate algorithm~\cite{kingma2015method} using the learning rate of $8\times10^{-4}$ and hidden layers with 1000 units. 

We train a model with pointer softmax where we assign pointers for the rarest 60 words and 
the rest of the words are predicted from the shortlist softmax of size 540. We observe that 
increasing the inverse temperature of the sigmoid output of $d_t$ to 2, in other words 
making the decisions of $d_t$ to become sharper, works better, i.e. $d_t = \sigmoid(2x)$.

At the end of training with pointer softmax we obtain the error rate of $17.4\%$ and by using
softmax over all 600 tokens, we obtain the error-rate of $48.2\%$.

%\tred{Complete this part.}

\subsection{Summarization}

In these series of experiments, we use the annotated Gigaword corpus as described in 
\cite{namas}. Moreover, we use the scripts that are made available by the authors of \cite{namas} ~\footnote{https://github.com/facebook/NAMAS} to preprocess the data, which results to approximately 3.8M training examples. This script generates about 400K validation and an equal number of test examples, but we use a randomly sampled subset of $2000$ examples each for validation and testing. We also have made small modifications to the script to extract not only the tokenized words, but also system-generated named-entity tags. We have created two different versions of training data for pointers, which we call \textit{UNK}-pointers data and entity-pointers data respectively.
 
For the \textit{UNK}-pointers data, we trim the vocabulary of the source and target data in 
the training set and replace a word by the UNK token whenever a word occurs less than 5 
times in either source or target data separately. Then, we create pointers from each UNK 
token in the target data to the position in the corresponding source document where the same word occurs in the source, as seen in the data before UNK's were created. It is possible that the source can have an UNK in the matching position, but we still created a pointer in this scenario as well. The resulting data has 2.7 pointers per 100 examples in the training set and 9.1 pointers rate in the validation set.
 
In the entity-pointers data, we exploit the named-entity tags in the annotated 
corpus and first anonymize the entities by replacing them with an integer-id that 
always starts from 1 for each document and increments from left to right. Entities 
that occur more than once in a single document share the same id. We create the 
anonymization at token-level, so as to allow partial entity matches between the source and 
target for multi-token entities. Next, we create a pointer from the target to source 
on similar lines as before, but only for exact matches of the anonymized entities. 
The resulting data has 161 pointers per 100 examples in the training set and 139 
pointers per 100 examples in the validation set.
 
If there are multiple matches in the source, either in the UNK-pointers data or the
entity-pointers data, we resolve the conflict in favor of the first occurrence of 
the matching word in the source document. In the UNK data, we model the UNK tokens 
on the source side using a single placeholder embedding that is shared across all documents, and in the entity-pointers data, we model each entity-id in the source by a distinct placeholder, each of which is shared across all documents.
 
In all our experiments, we use a bidirectional GRU-RNN \cite{gru_rnn} for the encoder and a uni-directional RNN for the decoder. To speed-up training, we use the large-vocabulary trick \cite{Jean2014} where we limit the  vocabulary of the softmax layer of the decoder to $2000$ words dynamically chosen from the words in the source documents of each batch and the most common words in the target vocabulary. In both experiments, we fix the embedding size to $100$ and the hidden state dimension to $200$. We use pre-trained word2vec vectors trained on the same corpus to initialize the embeddings, but we finetune them by backpropagating through the pre-trained embeddings during training. Our vocabulary sizes are fixed to 125K for source and 75K for target for both experiments.

We use the reference data for pointers for the model only at the training time. During the test time, the switch makes a decision at every timestep on which softmax layer to use.
 
For evaluation, we use full-length Rouge F1 using the official evaluation tool \footnote{\url{http://www.berouge.com/Pages/default.aspx}}. In their work, the authors of \cite{nmt} use full-length Rouge Recall on this corpus, since the maximum length of limited-length version of Rouge recall of 75 bytes (intended for DUC data) is already long for Gigaword summaries. However, since full-length Recall can unfairly reward longer summaries, we also use full-length F1 in our experiments for a fair comparison between our models, independent of the summary length.
 
The experimental results comparing the Pointer Softmax with NMT model are displayed in Table \ref{tbl:gigaword_summarization_unk} for the \textit{UNK} pointers data and in Table \ref{tbl:gigaword_summarization} for the entity pointers data. As our experiments show, pointer softmax improves over the baseline NMT on both UNK data and entities data. Our hope was that the improvement would be larger for the entities data since the incidence of pointers was much greater. However, it turns out this is not the case, and we suspect the main reason is anonymization of entities which removed data-sparsity by converting all entities to integer-ids that are shared across all documents. We believe that on de-anonymized data, our model could help more, since the issue of data-sparsity is more acute in this case. 
%These experiments show that adding PS can improve the performance of the model over the large-vocabulary trick, however the improvements when placeholders are used is much less. We used $1024$ hidden units for both encoder and the decoder GRU.

\begin{table}[ht]
\centering
\caption{Results on Gigaword Corpus when pointers are used for UNKs in the training data, using Rouge-F1 as the evaluation metric.}
\label{tbl:gigaword_summarization_unk}
\begin{tabular}{@{}llll@{}}
\toprule
%\multicolumn{4}{l}{\centering{Modelling UNK's using pointers}} \\ \midrule
                 & Rouge-1   & Rouge-2  & Rouge-L  \\
NMT + lvt             & 34.87     & 16.54    & 32.27    \\
NMT + lvt + PS   & \textbf{35.19}     & \textbf{16.66}    & \textbf{32.51}    \\ \bottomrule
\end{tabular}
\end{table}

\begin{table}[ht]
\centering
\caption{Results on anonymized Gigaword Corpus when pointers are used for entities, using Rouge-F1 as the evaluation metric.}
\label{tbl:gigaword_summarization}
\begin{tabular}{@{}llll@{}}
\toprule
%\multicolumn{4}{l}{\centering{Modelling UNK's using pointers}} \\ \midrule
                 & Rouge-1   & Rouge-2  & Rouge-L  \\
NMT + lvt              & 34.89     & \textbf{16.78}    & 32.37    \\
NMT + lvt + PS   & \textbf{35.11}     & 16.76    & \textbf{32.55}    \\ \bottomrule
\end{tabular}
\end{table}

\begin{table}[ht]
\centering
\caption{Results on Gigaword Corpus for modeling UNK's with pointers in terms of recall.}
\label{tbl:gigaword_summarization_unk_recall}
\begin{tabular}{@{}llll@{}}
\toprule
%\multicolumn{4}{l}{\centering{Modelling UNK's using pointers}} \\ \midrule
                 & Rouge-1   & Rouge-2  & Rouge-L  \\
NMT + lvt              & 36.45     & 17.41    & 33.90    \\
NMT + lvt + PS   & \textbf{37.29}     & \textbf{17.75}    & \textbf{34.70}    \\ \bottomrule
%\cite{namas} & 31.47 & 12.73 & 28.54 \\
\end{tabular}
\end{table}

In Table \ref{tbl:gigaword_summarization_unk_recall}, we provide the results for summarization on Gigaword corpus in terms of recall as also similar comparison is done by \cite{namas}. We observe improvements on all the scores with the addition of pointer softmax. Let us note that, since the test set of \cite{namas} is not publicly available, we sample $2000$ texts with their summaries without replacement from the validation set and used those examples as our test set.

In Table \ref{tbl:ex_summaries} we present a few system generated summaries from the Pointer Softmax model trained on the \textit{UNK} pointers data. From those examples, it is 
apparent that the model has learned to accurately point to the source positions whenever 
it needs to generate rare words in the summary.

\begin{table*}[ht]
\centering
\caption{Generated summaries from NMT with PS. Boldface words are the words copied from the source.}
\label{tbl:ex_summaries}

\begin{tabular}{|p{4cm}|p{10cm}|}
\hline
\textbf{Source \#1}         & china 's tang \textbf{gonghong} set a world record with a clean and jerk lift of \#\#\# kilograms to win the women 's over-\#\# kilogram weightlifting title at the asian games on tuesday .                                  \\ \hline
\multicolumn{1}{|p{4cm}|}{\textbf{Target \#1}}         & \multicolumn{1}{p{10cm}|}{china 's tang \textless unk\textgreater,sets world weightlifting record}                                                                                                                          \\ \hline
\multicolumn{1}{|p{4cm}|}{\textbf{NMT+PS \#1}} & \multicolumn{1}{p{10cm}|}{china 's tang \textbf{gonghong} wins women 's weightlifting weightlifting title at asian games}                                                                                                           \\ \hline
\multicolumn{1}{|p{4cm}|}{\textbf{Source \#2}}         & \multicolumn{1}{p{10cm}|}{owing to criticism , nbc said on wednesday that it was ending a \textbf{three-month-old} experiment that would have brought the first liquor advertisements onto national broadcast network television .} \\ \hline
\multicolumn{1}{|p{4cm}|}{\textbf{Target \#2}}         & \multicolumn{1}{p{10cm}|}{advertising : nbc retreats from liquor commercials}                                                                                                                                              \\ \hline
\multicolumn{1}{|p{4cm}|}{\textbf{NMT+PS \#2}} & \multicolumn{1}{p{10cm}|}{nbc says it is ending a \textbf{three-month-old} experiment}                                                                                                                                              \\ \hline
\multicolumn{1}{|p{4cm}|}{\textbf{Source \#3}}         & \multicolumn{1}{p{10cm}|}{a senior trade union official here wednesday called on ghana 's government to be `` \textbf{mindful} of the plight '' of the ordinary people in the country in its decisions on tax increases .}          \\ \hline
\multicolumn{1}{|p{4cm}|}{\textbf{Target \#3}}         & \multicolumn{1}{|p{10cm}|}{tuc official,on behalf of ordinary ghanaians}                                                                                                                                                    \\ \hline
\textbf{NMT+PS \#3}                       & ghana 's government urged to be \textbf{mindful} of the plight                                                                                                                                                                \\ \hline
\end{tabular}
\end{table*}
\subsection{Neural Machine Translation}
In our neural machine translation (NMT) experiments, we train NMT models with attention
over the Europarl corpus \cite{nmt} over the sequences of length up to $50$ for English to 
French translation. \footnote{In our experiments, we use an existing code, provided in \url{https://github.com/kyunghyuncho/dl4mt-material}, 
and on the original model we only changed the last softmax layer for our experiments}. 
All models are trained with early-stopping which is done based on the negative log-likelihood (NLL) on the development set. Our evaluations to report the performance of our models are done on {\tt newstest2011} by using BLUE score.~\footnote{We compute the BLEU score using the multi-blue.perl script from Moses on tokenized sentence pairs.}

We use $30,000$ tokens for both the source and the target language shortlist vocabularies~($1$ 
of the token is still reserved for the unknown words). The whole corpus contains $134,831$
unique English words and $153,083$ unique French words. We have created a word-level 
dictionary from French to English which contains translation of 15,953 words that 
are neither in shortlist vocabulary nor dictionary of common words for both the source 
and the target. There are about $49,490$ words shared between English and French parallel corpora of Europarl. 

During the training, in order to decide whether to pick a word from the source sentence using 
attention/pointers or to predict the word from the short-list vocabulary, we use the following simple heuristic. If the word is not in the short-list vocabulary, we first check if the same word 
$\vy_t$ appears in the source sentence. If it is not, we then check if a translated version of the word exists 
in the source sentence by using a look-up table between the source and the target language. If the word is in the source sentence, we then use the location of the word in the source as the target. Otherwise we check if one of the English senses from the cross-language dictionary of the French 
word is in the source. If it is in the source sentence, then we use the location of that word as our translation. Otherwise we just use the argmax of $\vl_t$ as the target.

For switching network $d_t$, we observed that using a two-layered MLP with noisy-tanh activation \cite{gulcehre2016noisy} function with residual connection from the lower layer \cite{he2015deep} activation function to the upper hidden layers improves the BLEU score about 1 points over the $d_t$ using ReLU activation function. We initialized the biases of the last sigmoid layer of $d_t$ to $-1$ such that if $d_t$ becomes more biased toward choosing the shortlist vocabulary at the beginning of the training. We renormalize the gradients if the norm of the gradients exceed $1$ \cite{pascanu2012difficulty}. 

\begin{table}[ht]
\centering
\caption{Europarl Dataset (EN-FR)}
\label{tbl:NMT_europarl}
\begin{tabular}{@{}ll@{}}
\toprule
%\multicolumn{4}{l}{\centering{Modelling UNK's using pointers}} \\ \midrule
                 & BLEU-4     \\
NMT              & 20.19         \\
NMT + PS   & \textbf{23.76}   \\ \bottomrule
\end{tabular}
\end{table}

In Table \ref{tbl:NMT_europarl}, we provided the result of NMT with pointer softmax and we
observe about $3.6$ BLEU score improvement over our baseline.

\begin{figure}[ht]
    \includegraphics[width=0.95\columnwidth]{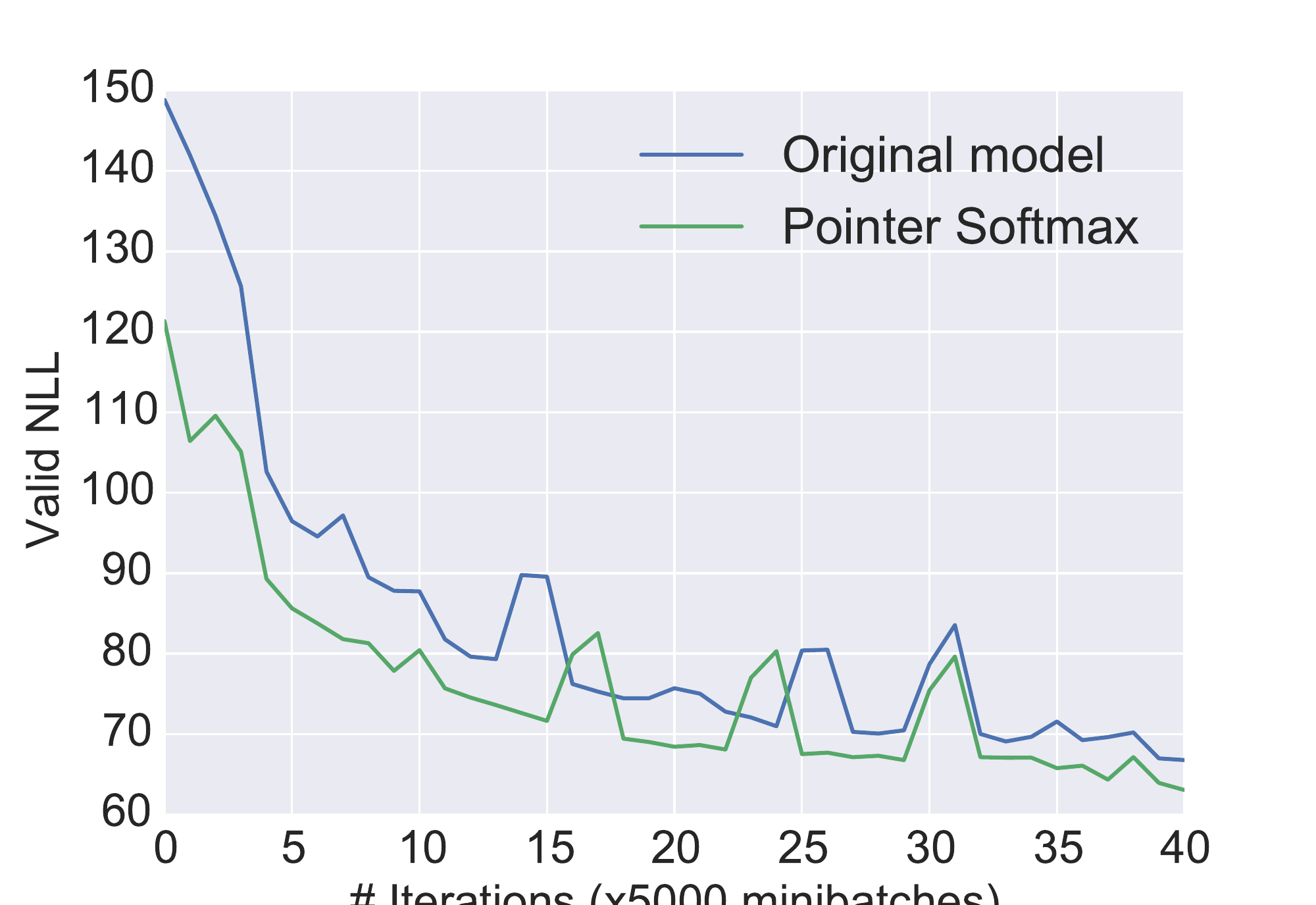}
    \caption{A comparison of the validation learning-curves of the same NMT model 
    trained with pointer softmax and the regular softmax layer. As can be seen from 
    the figures, the model trained with pointer softmax converges faster than the 
    regular softmax layer. Switching network for pointer softmax in this Figure uses 
    ReLU activation function.} 
    \label{fig:val_curves}
    \centering
\end{figure}

In Figure \ref{fig:val_curves}, we show the validation curves of the NMT model with attention and the NMT model with shortlist-softmax layer. Pointer softmax converges faster in terms of number of minibatch updates and achieves a lower validation negative-log-likelihood (NLL) ($63.91$) after $200k$ updates over the Europarl dataset than the NMT model with shortlist softmax trained for $400k$ minibatch updates~($65.26$). Pointer softmax converges faster than the model using the shortlist softmax, because the targets provided to the pointer softmax also acts like guiding hints to the attention.

\section{Conclusion}
%\tred{TODO: Figure out why references do not show up}
In this paper, we propose a simple extension to the traditional soft attention-based shortlist softmax by using pointers over the input sequence. We show that the whole model can be trained jointly with single objective function. We observe noticeable improvements over the baselines on machine translation and summarization tasks by using pointer softmax. By doing a very simple modification over the NMT, our model is able to generalize to the unseen words and can deal with rare-words more efficiently. For the summarization task on Gigaword dataset, the pointer softmax was able to improve the results even when it is used together with the large-vocabulary trick. In the case of neural machine translation, we observed that the training with the pointer softmax is also improved the convergence speed of the model as well. For French to English machine translation on Europarl corpora, we observe that using the pointer softmax can also improve the training convergence of the model.

\bibliography{acl2016}
\section*{Acknowledgments}
We would like to thank the developers of
Theano~\footnote{\url{http://deeplearning.net/software/theano/}}, for developing 
such a powerful tool for scientific computing \cite{2016arXiv160502688short}. We acknowledge the support of the following organizations for research funding and computing support: NSERC, Samsung, Calcul Qu\'{e}bec, Compute Canada, the Canada Research Chairs and
CIFAR. C. G. thanks for IBM T.J. Watson Research for funding this research during 
his internship between October 2015 and January 2016.

\bibliographystyle{acl2016}
\end{document}